\documentclass[sn-mathphys,Numbered]{sn-jnl}%

\usepackage{bbm}%
\usepackage{graphicx}%
\usepackage{multirow}%
\usepackage{amsmath,amssymb,amsfonts}%
\usepackage{amsthm}%
\usepackage{mathrsfs}%
\usepackage[title]{appendix}%
\usepackage{xcolor}%
\usepackage{textcomp}%
\usepackage{manyfoot}%
\usepackage{booktabs}%
\usepackage{algorithm}%
\usepackage{algorithmicx}%
\usepackage{algpseudocode}%
\usepackage{listings}%

\theoremstyle{thmstyleone}%

\theoremstyle{thmstyletwo}%

\theoremstyle{thmstylethree}%

\begin{document}

\title[Article Title]{Transformable Gaussian Reward Function for Socially-Aware Navigation with Deep Reinforcement Learning}

\author[1]{\fnm{Jinyeob} \sur{Kim}}\email{wls2074@khu.ac.kr}

\author[2]{\fnm{Sumin} \sur{Kang}}\email{suminsk@khu.ac.kr}

\author[2]{\fnm{Sungwoo} \sur{Yang}}\email{p1112007@khu.ac.kr}

\author[1]{\fnm{Beomjoon} \sur{Kim}}\email{1222kbj@khu.ac.kr}

\author*[2]{\fnm{Yura} \sur{Jargalbaatar}}\email{jargalbaatar@khu.ac.kr}

\author*[2]{\fnm{Donghan} \sur{Kim}}\email{donghani@khu.ac.kr}

\affil[1]{\orgdiv{Department of Artificial Intelligence}, \orgname{College of Software, Kyung Hee University}, \city{Yongin}, \country{Republic of Korea}}

\affil[2]{\orgdiv{Department of Electronic Engineering (AgeTech-Service Convergence Major)}, \orgname{College of Electronics \& Information, Kyung Hee University}, \city{Yongin}, \country{Republic of Korea}}

\abstract{Robot navigation has transitioned from prioritizing obstacle avoidance to adopting socially-aware navigation strategies that accommodate human presence. Consequently, socially-aware navigation in dynamic human-centric environments has gained prominence in the field of robotics. Although reinforcement learning techniques have fostered the advancement of socially-aware navigation, defining appropriate reward functions, particularly in congested environments, poses a significant challenge. These rewards, crucial for guiding robot actions, demand intricate human-crafted design owing to their complex nature and inability to be set automatically. The multitude of manually designed rewards poses issues such as hyperparameter redundancy, imbalance, and inadequate representation of unique object characteristics. To address these challenges, we introduce a transformable Gaussian reward function (TGRF). The TGRF significantly reduces the burden of hyperparameter tuning, displays adaptability across various reward functions, and demonstrates accelerated learning rates, particularly in crowded environments utilizing deep reinforcement learning (DRL). We introduce and validate the TGRF in sections highlighting its conceptual background, characteristics, experiments, and real-world applications, paving the way for a more effective and adaptable approach to robotics. The complete source code is available at https://github.com/JinnnK/TGRF.}


\maketitle
\section{Introduction}\label{sec1}

Over the years, robotics has shown a persistent interest in robot navigation. Initially, research focused on basic obstacle avoidance and random navigation strategies [1-3]. Advances in navigation techniques have led to simultaneous localization and mapping (SLAM) [4-6], wherein robots estimate their positions and create maps for effective movement. Strategies have expanded to address dynamic environments [7-17], as robotics has consistently pursued advancements in navigation.

However, despite the integration of robots into human environments, the effective avoidance of collisions remains a significant challenge. While safety, performance, and navigation in static environments have been emphasized [18], the field has shifted its focus to socially-aware navigation, which is recognized as crucial [12-14, 16, 17]. Socially-aware navigation integrates perception, intelligence, and behavior to adhere to social norms [19, 20], necessitating the critical ability to differentiate between humans and static objects.

\begin{figure}[h]%
\centering
\includegraphics[width=0.9\textwidth]{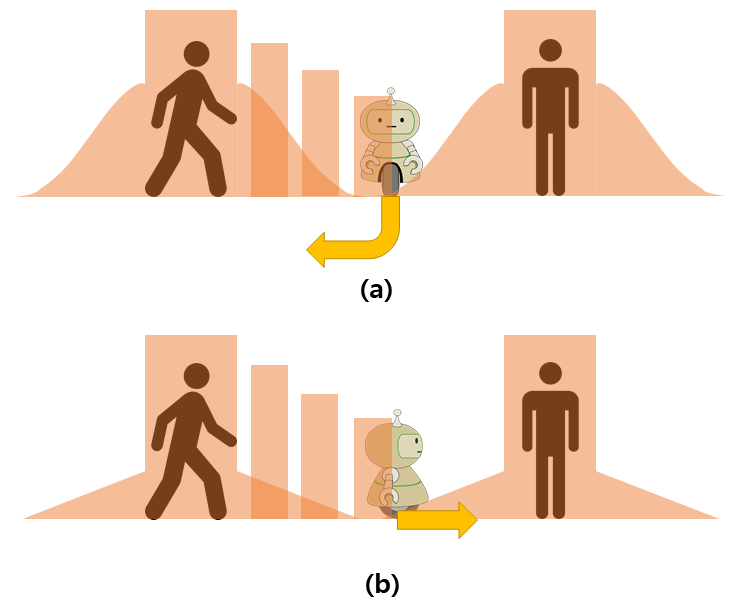}
\caption{Robot's actions when adequate reward functions are used or not. Red shapes represent penalties, and yellow arrows indicate robot's actions. Penalties are imposed when the robot is in proximity to humans, within their surroundings, or moving in their direction.}\label{fig1.png}
\end{figure}

To address this challenge, two main research directions have emerged: reactive navigation [9-11, 15] and navigation utilizing reinforcement learning (RL) [21-24]. Reactive navigation responds to real-time sensor data, with limitations in predicting future movements. RL employs the Markov decision process (MDP) and deep reinforcement learning (DRL) [12-14, 16, 17, 25] to leverage deep neural networks for well-informed decisions and enable robots to navigate safely in human environments.

However, the challenges in defining these reward functions become particularly evident in crowded environments where robots navigate [21, 26-28]. These reward functions essentially serve as the guiding principles for steering the actions of agents by evaluating the potential value of each action. As a result, human-crafted rewards have become indispensable because they cannot be set automatically. However, as demonstrated in Fig. 1, inadequately designed reward functions can induce risky behaviors in human–robot interactions. Moreover, the manual design of numerous rewards presents several critical issues.

First, the proliferation of distinct reward functions for various components in crowded environments necessitates a redundant number of hyperparameters [12-14, 16, 17, 25, 29-31]. Each reward demands tailored functions that align with its specific attributes such as distance from humans, direction toward the goal, or even human intentions. This surplus of reward functions results in excessive fine-tuning of the hyperparameters, leading to imbalances in the rewards. These imbalances can inadvertently steer robots toward humans, thereby increasing the risk of collisions [12]. Such disparities in action priorities caused by reward imbalances pose a significant challenge in tuning hyperparameters and identifying inadequate rewards.

Second, the fixed form of each reward function is neither temporally nor experimentally efficient. Static context-specific reward functions were used [12, 14, 16, 17, 32-35]. However, they often fail to adequately represent their unique characteristics. Even when the same formula is employed, diverse attributes may not be captured accurately. This discrepancy requires extensive empirical design and experimentation with specific reward functions to achieve higher performance.

Third, the hurdles can be extended to effective learning [27]. Crafting appropriate reward functions with the correct number of hyperparameters remains a significant challenge, which leads to collisions and hinders robot learning.

This paper proposes a transformable Gaussian reward function (TGRF) that aims to guide robots along safer routes. This approach makes several crucial contributions to the literature. (1) A smaller number of hyperparameters significantly alleviates the burden of parameter tuning, thus expediting the search for the optimal reward function. (2) The TGRF demonstrated adaptability to various reward functions through dynamic shape adjustments. Such adaptability is in stark contrast to previous models [13], which often require extensive redesigns for shape changes. (3) The TGRF exhibits accelerated learning rates, notably in crowded environments, effectively harnessing the potential of DRL.

To demonstrate the performance of the TGRF, we introduce the key points in reward shaping and relevant papers for comparison in the experiments in Section II. In Section III, we present background knowledge and characteristics of the TGRF and introduce the reward functions using the TGRF. In Section IV, we present two experiments conducted to demonstrate the performance of the TGRF and present the results of this study's application in real environments; finally, we conclude the paper in Section VI. 

\section{Related Works}\label{sec2}

\subsection{Integration of Prior Knowledge through Human-Delivered Reward Functions}\label{subsec2}

RL is a machine learning approach operating within the MDP [21], where an agent interacts with a specific environment and receives feedback in the form of rewards. The primary objective is to achieve the maximum cumulative reward that drives policy learning. The reward function significantly influences the agent's decision-making process, steering it toward convergence to optimal policies.

However, a prevalent issue arises because agents typically lack complete information regarding all aspects of the environment [36]. In vast state spaces, the transitions between states and rewards may be unknown or stochastic. Moreover, the agent remains unaware of the consequences and outcomes of the actions until they are executed.

Consequently, in such scenarios, agents require substantial experience to converge to optimal policies for complex tasks in the absence of prior knowledge. To address this challenge, research on RL has explored reward shaping, aiming to guide agents toward making better decisions at appropriate times using suitable reward values [27, 28]. This approach aims to significantly reduce the learning time by fostering convergence to optimal policies without explicit prior knowledge.

Previous studies extensively explored incorporating prior knowledge into reward functions [26, 28, 37-40]. However, crafting reward functions encompassing general prior knowledge, such as the apprehension of collision risks based on proximity to humans or progress relative to the destination, is challenging because of various environmental and psychological factors. These factors render it impossible to express knowledge simply through mathematical formulations.

To address this challenge, recent research has focused on utilizing inverse reinforcement learning (IRL), wherein humans intervene at the intermediate stages to provide rewards [41]. In addition, a study utilizing natural language to communicate intermediate reward functions with agents has emerged [42]. These studies involved humans evaluating the actions of robots as rewards and assigning them accordingly. They demonstrated the transmission of rewards imbued with prior knowledge to agents during learning, thereby accelerating the learning process and enhancing algorithm performance.

However, because of their reliance on human intervention, these approaches are unsuitable for environments that require extensive learning or complex tasks without direct human involvement. Therefore, there is a growing need for research on reward shaping that considers prior knowledge and delivers high performance without direct human intervention, particularly in environments that demand substantial learning.

\subsection{Reward Function Analysis for Human Avoidance in Robot Navigation}\label{subsec2}

Recent studies have actively explored methods for robot navigation in environments where humans and robots coexist [12-14, 16, 17, 25, 29-31]. Their reward functions commonly employ different formulas based on objectives without direct human intervention and can be broadly classified into four types. These include rewards categorized as follows: reaching the destination, $r_{goal}(s_t)$; collision avoidance with humans, $r_{col}(s_t )$; distance from humans, $r_{disc}(s_t )$; and distance from the destination, $r_{pot}(s_t )$.

$r_{goal}(s_t)$ and $r_{col}(s_t)$ typically assume consistent values, whereas $r_{disc}(s_t)$ and $r_{pot}(s_t)$ encompass numerous formulas that reflect prior knowledge, such as linear, L2 norm, or exponential functions. For instance, $r_{dist}(s_t)$ consistently imposes a larger penalty as the distance between humans and robots diminishes. This design aligns with the psychological theory of proxemics [20], which evaluates discomfort based on interpersonal distances, and integrates prior knowledge about the potential discomfort associated with varying distances between humans and robots.

In addition, $r_{pot}(s_t)$ incentivizes the robot's faster arrival at the destination by applying rewards or penalties based on changes in the L2 norm distance between the robot and the destination. These approaches reflect rational strategies by integrating prior knowledge of discomfort levels associated with distances (proxemics) and apprehensions regarding collision probabilities based on proximity to humans. Moreover, they encourage robots to expedite their arrival at their destination through rewards/penalties based on the changing distance from the destination, using the L2 norm.

However, studies related to reward shaping and RL argue that it is crucial to verify whether rewards take appropriate forms and maintain suitable proportions [27, 28]. If the shapes of the rewards are inadequate for the objectives or overly biased, the robot may steer its learning process in a direction not intended by the algorithm, potentially leading to the freezing robot problem [42]. In addition, an excessive number of hyperparameters may hinder the search for optimal performance.

The aforementioned studies experimentally determined the reward functions and counts of the hyperparameters. As a result, some were excessively simplistic, preventing researchers from intuitively adjusting rewards through hyperparameters, whereas others exhibited complex structures that hindered the straightforward modification of hyperparameters. This resulted in significant time consumption to achieve optimal performance and limitations in adjusting inadequate rewards, necessitating a redesign of the reward function.

For instance, in [13], researchers designed simple reward functions with redundant hyperparameters. This resulted in a substantial nine-fold difference between $r_{pred}(s_t)$ and $r_{disc}(s_t)$. This leads to situations in which the robot favors actions with smaller penalties from $r_{disc}(s_t)$ over larger penalties from $r_{pred}(s_t)$, thereby resulting in intrusion and collision.

This directly affects the learning process, rendering the task of identifying the appropriate reward function and hyperparameters more challenging, and requiring formula modification.

Therefore, this paper proposes the TGRF, which allows for intuitive and versatile applications with fewer hyperparameters. Enabling researchers to adjust rewards intuitively reduces the time required to explore suitable reward functions and ensures optimal performance by finely tuning reward balances. To substantiate this claim, we directly compared the reward function used in socially-aware navigation (SCAN) [16], decentralized structural-recurrent neural network (RNN) (DS-RNN) [17], Gumbel social transformer + human–human attention (GST+HH Attn) [13, 43], and crowd-aware memory-based RL (CAM-RL) [31].

\section{Suggested Reward Function}\label{sec3}

In Section III-A, we briefly introduce background knowledge regarding the model. In Section III-B, we elaborate on the proposed TGRF. Finally, in Section III-B-3, we describe the application of the TGRF to the reward functions within the environment and model of [13].

\subsection{Preliminaries}\label{subsec3}

\subsubsection{Markov decision process (MDP) and navigation methods}\label{subsubsec3}

\begin{figure}[h]%
\centering
\includegraphics[width=0.5\textwidth]{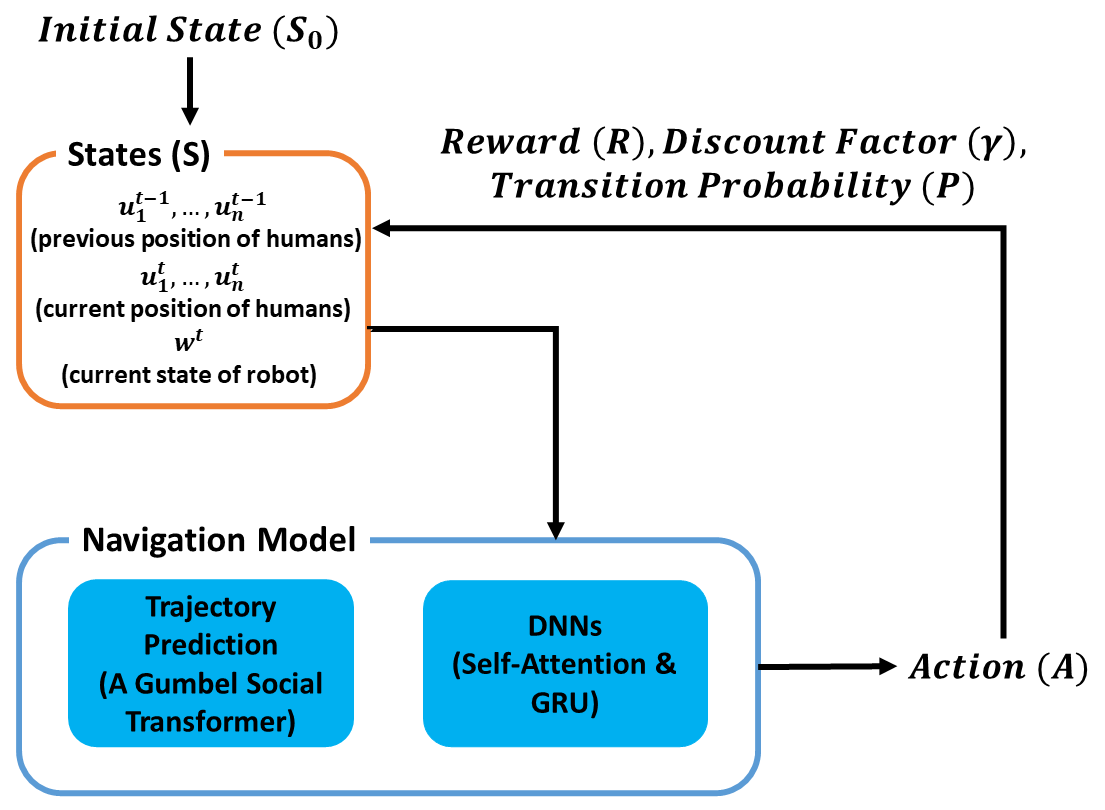}
\caption{\centering Diagram of an MDP with navigation model applied}\label{fig2.png}
\end{figure}

In the crowded environment described by the tuple ⟨S,A,P,R,$\gamma$,$S_{0}$⟩, robots and humans follow distinct policies during simultaneous movements. It includes states (S), actions (A), rewards (R), transition probabilities (P), discount factor ($\gamma$), and initial state ($S_{0}$) [21].

The robot utilizes a learning-based model called GST + HH Attn [13], which employs an attention mechanism [45] for navigation based on the past and current positions of humans to predict future positions.

In the GST + HH Attn model, the robot begins at $S_{0}$ and receives as state information: the robot's current position ($p_{x}$,$p_{y}$), velocity ($v_{x}$,$v_{y}$), destination ($g_{x}$,$g_{y}$), maximum velocity $v_{max}$  , angle $\theta$, and robot radius $\rho$, along with the set of $i-th$ person's current position information ($p_x^i$,$p_y^i$) at time $t$, denoted as $u_i^t$. $i$ can vary from zero to the maximum number of people, $n_{max}$. However, owing to the limitations in the light detection and ranging (LiDAR) range, the total number of people n in the state received by the robot might differ from $n_{max}$ and can vary at each time t. In addition, the robot remembers the past positions of the detected people up to a constant M step using the trajectory-prediction algorithm (GST) to nonlinearly predict $\hat{u}_i^{t+1:t+K}$, the predicted positions of individuals from time t+1 to t+K.

Therefore, the GST + HH Attn model leverages the set of predicted positions $\hat{U}_i^{t+1:t+K}=[\hat{u}_0^{t+1:t+K},\hat{u}_1^{t+1:t+K},...,\hat{u}_n^{t+1:t+K}]$ up to constant K steps and the set of current positions $U^t=[u_0^t,u_1^t,u_2^t,…,u_n^t]$ of people to predict their overall dynamics through an attention mechanism. It employs a type of RNN called a GRU to detect and process long-term dependencies, thus enabling a broader field of view [46]. This model focuses on nearby individuals when selecting their actions. Furthermore, it receives feedback and updates its state for the next step based on the selected actions. In addition, this study employed five learning-based methods using the same model:

\begin{itemize}
\item DS-RNN: A model utilizing an RNN. However, it did not predict the trajectories.
\item No pred + HH Attn: Attention-based model excluding trajectory prediction ($r_{pred}=0$).
\item Const vel + HH Attn: The experimental case assumes that the trajectory-prediction algorithm predicts the trajectories to move at a constant velocity.
\item Truth + HH Attn: This experiment assumes that the robot predicts the actual human trajectory.
\item GST + HH Attn: Scenarios in which the robot predicts the human trajectory nonlinearly using GST.
\end{itemize}

For a detailed model explanation, please refer to [13, 17].

Each human was randomly assigned a starting location and destination. Humans exchange their location information $U^t=[u_0^t,u_1^t,u_2^t,…,u_n^t]$ with each other and calculate velocities based on their positions, engaging in collision avoidance by altering their speed and direction using reactive-based methods such as ORCA and SF [9-11].

Similar to robots, humans are designated destinations that aim to reach or modify using a predetermined probability. Humans possess attributes such as size, speed, and maximum speed.

However, the humans were assumed to disregard the robot’s location for two primary reasons: First, in crowded environments, human reactions to robots lead to subtle movements resulting in negligible positional changes. Even rapid shifts are infeasible in densely populated settings. Second, encapsulating human dynamics within a mathematical model is challenging and can impede the learning process of robots. The diversity of the manners in which humans react to robots, along with the associated uncertainties, makes modeling impractical. Furthermore, even if such modeling were feasible, it would interfere with the robot’s learning. Consequently, humans either do not perceive the robot’s location or ignore it.

Therefore, humans perform subsequent actions based on their own characteristics and information about the cur-rent positions and velocities of others.

\subsection{Transformable Gaussian Reward Function (TGRF)}\label{subsec3}

\subsubsection{Formula and number of hyperparameters}\label{subsubsec3}

TGRF leverages the characteristics of a normal distribution [47]. A normal distribution offers the advantage of being able to transform into various shapes using only two hyperparameters. This flexibility enhances the adaptability of the model, allowing adjustments to fit diverse prior knowledge and applying them to reward functions. Furthermore, it alleviates the burden on researchers in tuning hyperparameters, aiding in the swift identification of appropriate hyperparameters within a short timeframe.

The formula for the normal distribution (N) is as follows:

\begin{figure}[h]%
\centering
\includegraphics[width=0.9\textwidth]{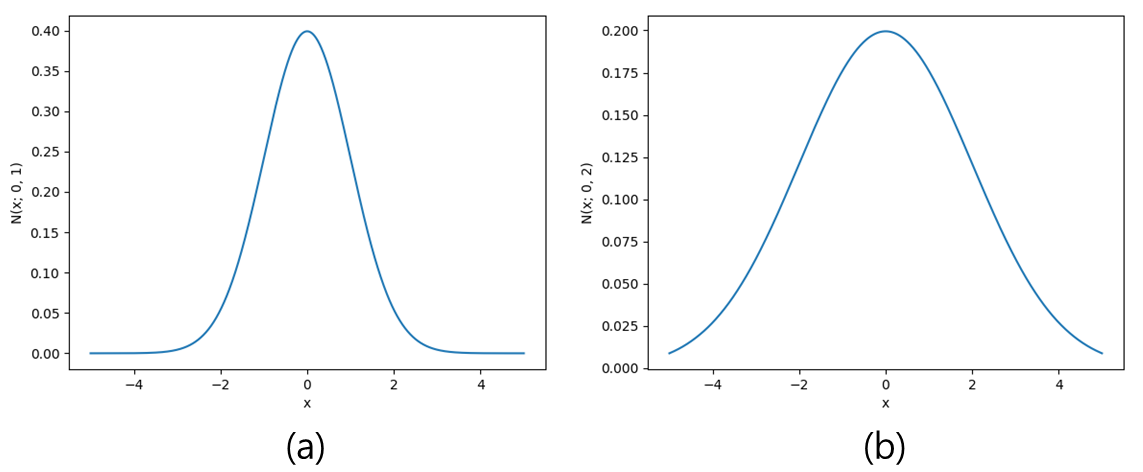}
\caption{Normal distribution. The X-axis denotes the X-value, and Y-axis represents the $N(x;\mu,\sigma)$. In (a), $\mu=0,\sigma=1$. In (b), $\mu=0,\sigma=2$.}\label{fig3.png}
\end{figure}

\begin{equation}
N(x; \mu, \sigma) = \frac{1}{\sqrt{2\pi\sigma}} \cdot \exp\left( -\frac{(x - \mu)^2}{2\sigma^2} \right)
.\label{eq1}
\end{equation}

The normal distribution involves variables x, mean $\mu$, and variance $\rho$. In Fig. 3, the normal distribution represents the shape in which the values of a random variable are distributed, exhibiting symmetry around $\mu$. It peaks at $x=\mu$ and decreases as x moves away from $\mu$. $\rho$ determines the width of the normal distribution, with the length of the width proportional to $\rho$. Therefore, the normal distribution has the advantage of being highly versatile in assuming various shapes with just two parameters, $\mu$ and $\rho$.

\begin{figure}[h]%
\centering
\includegraphics[width=0.9\textwidth]{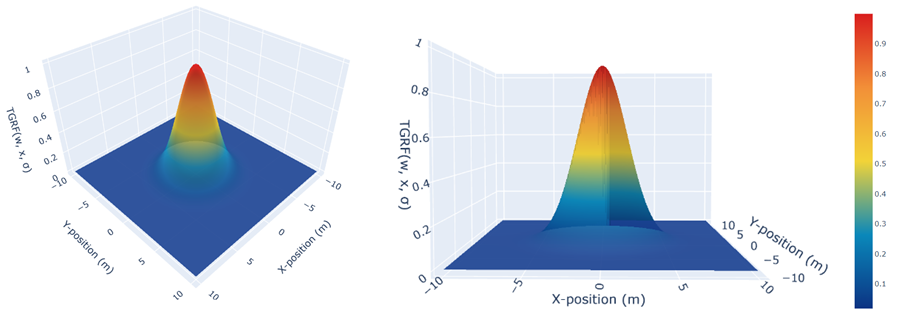}
\caption{TGRF. The X-axis denotes the X-position in meters, Y-axis represents the Y-position in meters, and Z-axis indicates TGRF value when  $w_{TGRF}=1$,$x_{TGRF}=dist(X,Y)$,$\mu_{TGRF}=0$,$\sigma_{TGRF}=2$.}\label{fig4.png}
\end{figure}

\begin{equation}
\begin{gathered}
TGRF(w_{TGRF},\mu_{TGRF},\sigma_{TGRF};x_{TGRF}) = \frac{w_{TGRF} \cdot N(x_{TGRF}; \mu_{TGRF}, \sigma_{TGRF})}{C_{\text{norm}}} \\
C_{\text{norm}} = \max_{x_{\text{norm}}} N(x_{\text{norm}}; \mu_{TGRF}, \sigma_{TRGF})
\end{gathered}
\end{equation}

TGRF leverages the characteristics of a normal distribution. In (2), TGRF involves three hyperparameters: $w_{TGRF}$, representing the weight of TGRF; $\mu_{TGRF}$, denoting the mean; and $\rho_{TGRF}$, indicating the variance.

It incorporates a single variable $x_{TGRF}$. $C_{norm}$ ensures that the TGRF attains a maximum value of 1, irrespective of $\rho_{TGRF}$. This allows the scaling of the TGRF solely by $w_{TGRF}$, enabling researchers to intuitively control its maximum value, preventing the freezing robot problem [43], and balancing it with other reward values to attain the desired algorithmic performance.

$\rho_{TGRF}$ determines the transformability of TGRF. As $\lim_{\sigma_{TGRF} \to \infty} \mathcal{N}(x_{TGRF}; \mu_{TGRF}, \sigma_{TGRF})$, it takes on a constant form insensitive to changes in $x_{TGRF}$, while as $\lim_{\sigma_{TGRF} \to 0} \mathcal{N}(x_{TGRF}; \mu_{TGRF}, \sigma_{TGRF})$, it resembles an impulse function. This versatility enables the creation of diverse forms of TGRF, such as constant, linear, non-linear, Gaussian, etc., adaptable to specific objectives.

$\mu_{TGRF}$ determines the position of TGRF's maximum value. Thus, allowing adjustments of the maximum value position according to the objectives requiring specific values to be reached. The variable $x_{TGRF}$ signifies the variable in N($x_{TGRF}$; $\mu_{TGRF}$,$\rho_{TGRF}$). As a result, it ultimately represents a shape similar to Fig. 4.

\subsubsection{Transformability}\label{subsubsec3}

\begin{figure}[h]%
\centering
\includegraphics[width=0.9\textwidth]{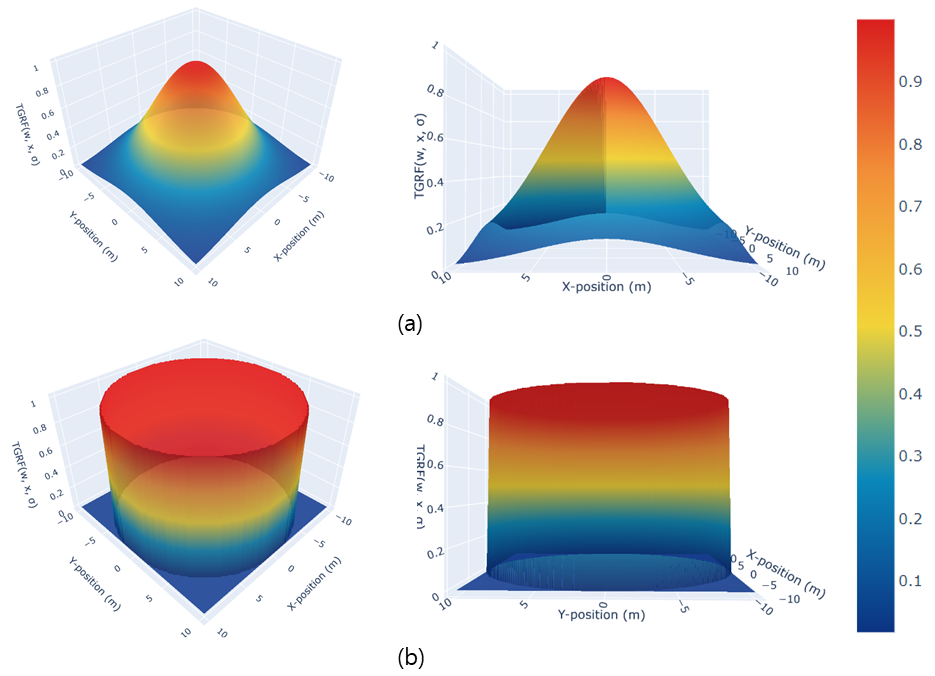}
\caption{Transformability of TGRF. The X-axis denotes the X-position in meters, Y-axis represents the Y-position in meters, and Z-axis indicates TGRF value. In (a), $w_{TGRF}=1$,$x_{TGRF}=dist(X,Y)$,$\mu_{TGRF}=0$,$\sigma_{TGRF}=5$. In (b), $w_{TGRF}=1$,$x_{TGRF}=dist(X,Y)$,$\mu_{TGRF}=0$,$\sigma_{TGRF}=5000$.}\label{fig5.png}
\end{figure}

The reward functions described in Section II-B have different formulas and limited flexibility. These constraints lead to complexity in the reward functions and difficulty in determining optimal performance. Therefore, researchers must redesign and tune them to achieve better performance. However, the TGRF offers versatility in generating various shapes.

Fig. 5 illustrates the creation of different shapes using the same TGRF by simply adjusting $\rho_{TGRF}$. Fig. 5a shows a TGRF that generates a continuous Gaussian distribution, making it suitable for moving objects or humans by applying varying penalties. Fig. 5b shows the discrete column-like shape. This configuration is suitable for stationary reward functions and objects. This approach allows various shapes via adjusting only a hyperparameter ($\rho_{TGRF}$) without requiring an additional function. This significantly reduces researchers’ time and effort while enabling fine-tuning to match the specific characteristics of objects.

\subsubsection{Reward function}\label{subsubsec3}

Reward $r(s_t,a_t)$ is categorized into five types. First, $r_{goal} (s_t)$ = 10 represents the reward when the robot successfully reaches its destination. Second, $r_{col} (s_t)$ = -10 serves as a penalty incurred upon colliding with another individual. Third, $r_{disc} (s_t)$ represents the penalty for entering a danger zone. Fourth, $r_{pot} (s_t)$ corresponds to the reward/penalty contingent on the change in distance to the destination $S_{goal}$. Finally, $r_{pred} (s_t)$ denotes the penalty invoked when entering a prediction.

\begin{figure}[h]%
\centering
\includegraphics[width=0.9\textwidth]{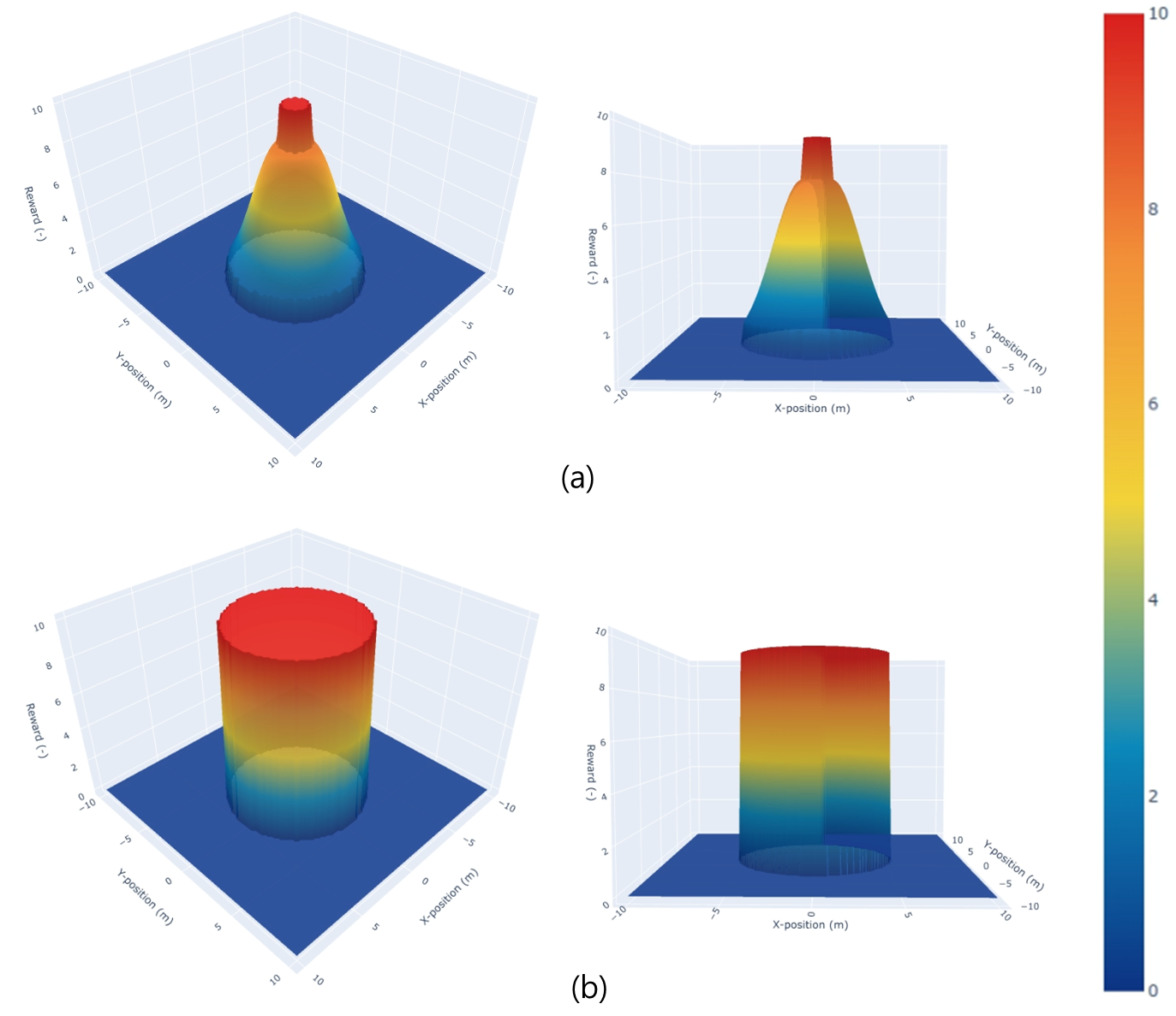}
\caption{TGRF applied to reward functions. The X-axis denotes the X-position in meters, Y-axis represents the Y-position in meters, and Z-axis indi-cates negative reward value. Central cylinder represents $r_{col}$, and surrounding distribution represents $r_{disc} (s_t)$. Beyond $d_{disc}$, $r_{disc} (s_t)$ becomes 0 ($s_t \notin S_{\text{danger zone}}
$). In (a), $w_{TGRF}=8$,$x_{TGRF}=dist(X,Y)$,$\mu_{TGRF}=0$,$\sigma_{TGRF}=3$. In (b), $w_{TGRF}=10$,$x_{TGRF}=dist(X,Y)$, $\mu_{TGRF}=0$, $\sigma_{TGRF}=1000$.}\label{fig6.png}
\end{figure}

The penalty $r_{disc} (s_t )$ is imposed when the robot enters the danger zone ($d_{min}$ is within $d_{disc}$) determined by the nearest human distance, denoted as $d_{min}$. The formula used is as follows:

\begin{equation}
\begin{gathered}
r_{\text{disc}}(s_t) = \text{TGRF}(w_{\text{disc}},0,\sigma_{\text{disc}};d_{\text{min}})
\end{gathered}
\end{equation}

In (3) and Fig. 6a, $r_{disc} (s_t )$ is designed to prevent collisions with humans, following a Gaussian distribution with $\mu_{disc}$ = 0, whereby smaller values of $d_{min}$ (closer proximity to humans) result in a higher penalty. This design aligns with prior knowledge and exhibits a plausible and realistic shape.

$r_{disc} (s_t )$ can be adjusted using only two hyperparameters. By tuning $\rho_{disc}$, researchers can regulate the breadth of the Gaussian penalty concerning the distance between humans and the robot. This enables the robot to react more sensitively or less sensitively to the distance from humans. $w_{disc}$ directly scales the reward function, establishing a linear correlation with $r_{disc} (s_t )$, thereby enabling adjustment of the overall reward balance to prioritize driving tasks.

The potential reward $r_{pot} (s_t)$ represents the reward associated with the potential field and is defined as follows:

\begin{equation}
\begin{gathered}
r_{\text{pot}}(s_t) = \Delta d \cdot TGRF(1.5, \mu_{\text{pot}}, \sigma_{\text{pot}}; \mu_{\text{pot}}) \\
\Delta d = (-d_{\text{goal}}^t + d_{\text{goal}}^{t-1})
\end{gathered}
\end{equation}

$r_{pot} (s_t )$ plays a crucial role in guiding a robot toward its destination and mitigating the freezing robot problem [21]. However, high values of $r_{pot} (s_t )$ can lead to increased collision rates in humans. Therefore, in Equation (4), by setting $x_{pot}=\mu_{pot}$, we aimed to maintain a constant TGRF regardless of $\mu_{pot}$  and $\rho_{pot}$. In addition, by applying $w_{pot}$ = 1.5, we ensured equal weighting for all $\Delta d$, aiming for uniformity across them.

Penalty $r_{pred} (s_t )$ is the reward for the prediction and defined as follows:

\begin{equation}
\begin{gathered}
r_{\text{pred}}^i(s_t) = \min_{k=1,\ldots,K} \left( \mathbbm{1}_i^{t+k} \frac{r_{\text{col}}}{2^k} \right) \\
r_{\text{pred}}(s_t) = \min_{i=1,\ldots,n} r_{\text{pred}}^i(s_t)
\end{gathered}
\end{equation}

$r_{pred} (s_t )$ is used only in models that employ trajectory predictions. $r_{pred} (s_t )$ denotes the penalty value when the robot is positioned along the trajectory of the $i-th$ person. $\mathbbm{1}_{i}^{t+k}$ indicates whether the robot is in the predicted position of the $i-th$ person at time t+k or not. Thus, $r_{pred} (s_t )$ takes the smallest penalty among all individual trajectory penalties that the robot takes. In our experiments, to demon-strate the performance enhancement even with different reward functions, we adopted $r_{pred} (s_t )$ used in the GST + HH Attn from [13]. The final definition of the reward function is as follows:

\begin{equation}
r(s_t,a_t) = 
\begin{cases}
+10, & \text{if } s_t \in S_{\text{goal}} \\
-10, & \text{if } s_t \in S_{\text{collision}} \\
r_{\text{pred}}(s_t) + r_{\text{disc}}(s_t),  & \text{if } s_t \in S_{\text{danger zone}} \\
r_{\text{pred}}(s_t) + r_{\text{pot}}(s_t), & \text{otherwise}
\end{cases}
\end{equation}

In summary, TGRF offers the distinct advantage of intuitively and efficiently modifying the reward function with fewer hyperparameters. This enables the robot to make rational decisions and reduces time-consuming fine-tuning tasks.

\section{Simulation Experiments}\label{sec4}

This section describes the experimental setup, procedure, and outcomes.

\subsection{Experimental environment}\label{subsec4}

\begin{figure}[h]%
\centering
\includegraphics[width=1.0\textwidth]{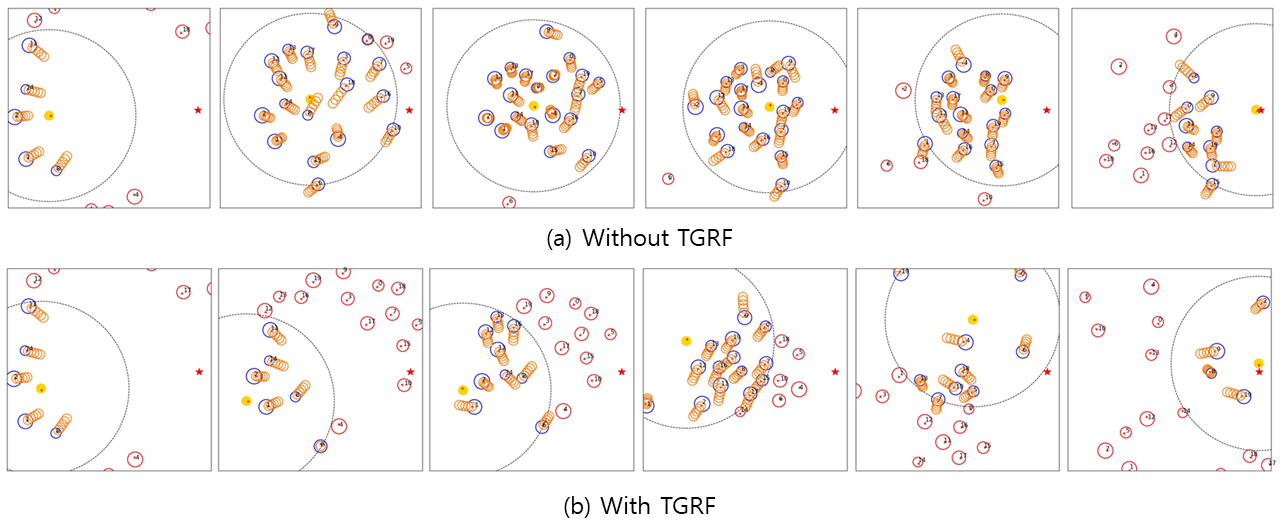}
\caption{Comparing scenarios with and without TGRF. Yellow circles represent robots, blue circles represent humans within the sensor range, red circles represent humans outside the sensor range, and orange circles in front of the blue circles indicate trajectories predicted by the GST.}\label{fig7.png}
\end{figure}

We employed a 2D environmental simulator as in previous studies [13]. This simulator features a 12 × 12m space, with a 360° field of view and 5\ m sensor range for LiDAR. A fixed number of humans (20) was used to represent a crowded setting.

\begin{equation}
\begin{aligned}
p_x[t+1] &= p_x[t] + v_x[t] \Delta t \\
p_y[t+1] &= p_y[t] + v_y[t] \Delta t
\end{aligned}
\end{equation}

Both humans and robots were operated using holonomic kinematics to determine their velocities $(a_t  = [v_x,v_y ]m/s)$ along the x- and y-axes. Holonomic kinematics refers to a state in which degrees of freedom can move independently without any constraints. This implies that robots and machines can move without limitations on their position or orientation. As a result, the action space of a robot is continuous, allowing both robots and humans to immediately achieve their desired speed within a time frame of $\Delta t$, assuming they operate within the maximum speed limit. Therefore, the positions of the humans and robots are continuously updated according to (7).

The robot has attributes such as a size of $\rho = 0.3m$ and a maximum speed of $v_{max}$  = $1.0m/s$. Humans also have characteristics such as a size ranging from 0.3 to 0.5m and a maximum speed varying between 0.5 and 1.5m/s. In addition, the locations and destinations of the robot were randomized, and the destinations were not set excessively close. If the robot collides with an individual, reaches its destination, or exceeds the maximum time T, the episode is terminated, leading to the beginning of a subsequent episode.

Random seeding was applied during training, resulting in varying outcomes for each training episode. To handle the varying outcomes, multiple training runs were conducted with a total time step of 2×$10^7$ for DS-RNN and 1×$10^7$ for the other algorithms. The learning rate was set to 4×$10^{-5}$ for all policies. Subsequently, test data were acquired from 500 test episodes. The evaluation metrics applied to the test data included the success rate (SR), average navigation time (NT) in seconds, path length (PL) in meters for successful episodes, and intrusion time ratio (ITR).

\subsection{Results}\label{subsec4}

\subsubsection{Results in Different Navigation Methods and Environment}\label{subsubsec4}

\begin{table}[h]
\caption{Navigation results using the reward function from [13] and TGRF. Humans follow ORCA.}\label{tab1}%
\begin{tabular}{@{}lllllll@{}}
\toprule
Reward & Navigation Method  & Mean(sigma) of SR & SR(\%) & NT(s) & PL(m) & ITR(\%) \\
\midrule
        & DS-RNN              & 35.5 (7.697)   & 44.0 & 20.48 & 20.58 & 15.45  \\
        & No pred + HH Attn   & 59.636 (4.848) & 67.0 & 17.49 & 20.30 & 17.22  \\
Without & Const vel + HH Attn & 65 (8.023)     & 81.0 & 17.34 & 21.95 & 6.15  \\
TGRF    & Truth + HH Attn     & 5.545 (1.616)  & 5.0  & 21.60 & 23.89 & 14.83  \\
        & GST + HH Attn       & 77.1 (5.718)   & 88.0 & 14.18 & 20.19 & 7.38  \\
\botrule
        & DS-RNN                  & 30.5 (4.843)     & 40.0  & 27.11 & 25.19 & 12.19  \\
With    & No pred + HH Attn       & 59.364 (6.692)   & 72.0  & 18.17 & 21.92 & 14.16  \\
TGRF    & Const vel + HH Attn     & 87.909 (3.029)   & 92.0  & 16.38 & 22.33 & 5.08  \\
(Ours)  & Truth + HH Attn         & 84.909 (4.776)   & 92.0  & 17.13 & 22.52 & 5.30  \\
        & GST + HH Attn           & 94.091 (2.843)   & 97.0  & 17.63 & 23.81 & 3.92  \\
\toprule
\end{tabular}
\end{table}

\begin{table}[h]
\caption{Navigation results using the reward function from [13] and TGRF. Humans follow SF.}\label{tab1}%
\begin{tabular}{@{}lllllll@{}}
\toprule
Reward & Navigation Method  & Mean(sigma) of SR & SR(\%) & NT(s) & PL(m) & ITR(\%) \\
\midrule
         & DS-RNN              & 29.8 (5.231)    & 36.0 & 23.26 & 27.13 & 13.38  \\
Without  & No pred + HH Attn   & 12.091 (8.062)  & 28.0 & 26.52 & 34.98 & 12.78  \\
TGRF     & Const vel + HH Attn & 92.182 (3.588)  & 96.0 & 14.74 & 21.49 & 5.24  \\
         & GST + HH Attn       & 91.636 (2.267)  & 95.0 & 13.74 & 20.47 & 5.37  \\
\botrule
        & DS-RNN               & 48.6 (9.013)     & 62.0  & 22.48 & 25.26 & 10.14  \\
With    & No pred + HH Attn    & 77.364 (6.079)   & 87.0  & 16.19 & 21.95 & 13.43  \\
TGRF    & Const vel + HH Attn  & 95.273 (1.911)   & 98.0  & 17.00 & 23.55 & 5.39  \\
(Ours)  & GST + HH Attn        & 92.909 (3.579)   & 96.0  & 15.37 & 21.91 & 5.81  \\
\toprule
\end{tabular}
\end{table}

We compared the performance with the rewards (from Section III-B-3) when the TGRF was applied to the performance without its application (reward function from [13]). The experiment involved the application of five different navigation methods to the robots, as described in Section III-A-1. Table I shows the performance when individuals adhere to ORCA, whereas Table II outlines the performance when adhering to SF. The hyperparameters are set to $w_{disc}=0.25$, $x_{disc}=d_{min}$, $\mu_{disc}=0$,$\rho_{disc}=0.2$, $d_{disc}=0.5$, $w_{pot}=1.5$,$x_{pot}=\mu_{pot}$, $\mu_{pot}=0$,$\rho_{pot}=1000$. From Tables I, and II and Fig. 7, we can identify three impacts of the TGRF.

First, the TGRF results in higher performance. As shown in Tables I and II, the TGRF significantly outperformed it in terms of SR. Notably, Table I shows that TGRF achieves an average SR of 94.091\%, whereas the reward function without TGRF attains only 77.1\% under the GST + HH Attn policy, marking a notable 17\% increase.

Hence, models employing the TGRF generally exhibit equal or superior performance in terms of SR and ITR. This implies that the TGRF demonstrates relatively high performance across various models. This signifies that the TGRF effectively incorporates prior knowledge based on the role of the reward function, indicating resilience in the freezing robot problem [43]. Consequently, it is evident that the robot demonstrates high SR by taking appropriate actions according to the situation.

Second, even when combined with different reward functions without the TGRF, the TGRF exhibits high performance. In Tables I and II, for models utilizing trajectory prediction, $r_{pred} (s_t)$ employs the reward function of a previous study without the TGRF. Nonetheless, it showed superior performance compared with [13], proving its adaptability to other reward functions.

Third, this leads to enhanced recognition of human intent and collision avoidance. Fig. 7a shows the behavior of the robot when a reward function without the TGRF was applied. The robot struggles when confronted with crowds. Notably, in the test cases, the robot ventured into the crowd, resulting in unintended collisions with humans while attempting to navigate the crowd. Similar situations were observed in the other test cases. This means that the robot selects aggressive or impolite behaviors, such as sidestepping, to avoid human and unintentional collisions, or making risky decisions to reach a destination faster, resulting in collisions. This behavior reflects a deficiency in understanding the broader intentions of humans and an imbalance in reward functions.

However, in Fig. 7b, the robot proactively positioned itself behind the crowd before converging at a single point. This means that the values of $r_{pred} (s_t )$, $r_{pot} (s_t )$, and $r_{disc} (s_t )$ related to human interactions were well balanced, enabling the robot to navigate effectively without colliding with individuals.

\begin{figure}[h]%
\centering
\includegraphics[width=1.0\textwidth]{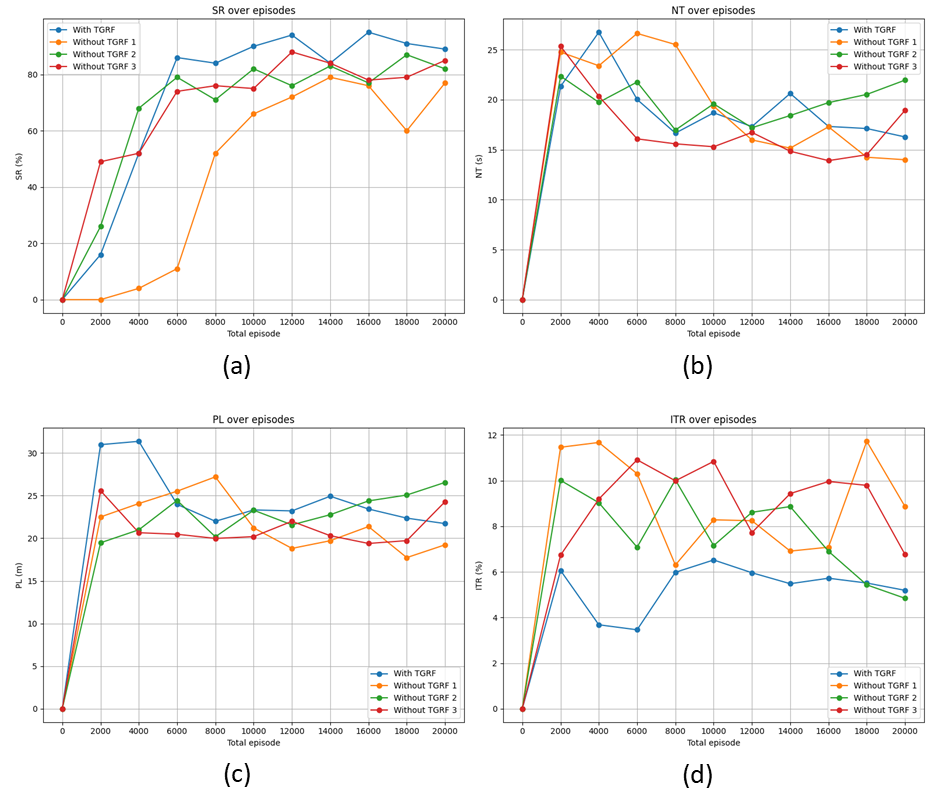}
\caption{Comparison of performance among four types of reward functions. Blue line represents the reward function incorporating TGRF, while or-ange line corresponds to the reward function in [13], green one reflects [16], and red signifies [31]. (a) denotes SR, (b) represents NT, (c) stands for PL, and (d) signifies ITR.}\label{fig8.png}
\end{figure}

This signifies that the TGRF effectively incorporates prior knowledge and that its priorities are well integrated into the policy. This suggests that the performance of the algorithm can be further enhanced when the TGRF is applied. Further evidence of this enhancement is reflected in the results in Tables I and II, where the average SR and standard deviation show similar or superior performances compared with previous iterations. In the other test cases, we observed that the robot selected a more secure and effective route than a faster and more dangerous route.

\subsubsection{Performance Comparison with Other Reward Functions}\label{subsubsec4}

In this experiment, we conducted trials applying the reward functions from previous studies [13, 16, 31] introduced in Section II-B along with the reward function incorporating the TGRF described in Section III-B-3. The navigation method employs the GST + HH Attn models. Participants adhered to the ORCA approach by recording SR, NT, PL, and ITR every 2000 episodes. The total episodes conducted were 20000 for GST + HH Attn.

First, it was observed that applying TGRF to the model led to an overall performance improvement compared with the other models. As shown in Fig. 8a, the reward function with the TGRF was able to drive the algorithm's performance up to a maximum of 95\% over 16000 total episodes. Conversely, the other reward functions achieved a maximum SR of 90\% This indicates that the TGRF harmonizes appropriately with the other reward functions, assuming a shape that aligns with the role of the rewards, thereby eliciting the algorithm's maximum performance.

As depicted in Fig. 8b, in the model applying the TGRF, the NT decreased with repeated learning, ultimately confirm-ing the second lowest NT. Correspondingly, in Fig. 8c, the second lowest PL was also observed. This is associated with the ITR, as lower NT and PL imply that the robot tolerates penalties owing to $r_{disc} (s_t)$ reaching the destination, resulting in a higher ITR. For instance, in the orange graph, the highest ITR, along with the low-est NT and PL values, can be observed. As a result, as shown in Fig. 7a, this leads to the choice of shorter and riskier paths, increasing the likelihood of not understanding human intentions and a higher possibility of collisions. However, as shown in Fig. 8d, the model incorporating the TGRF maintained the lowest ITR in most cases. This demonstrates that the TGRF selects the most efficient and safe paths compared to the other models, while maintaining the highest SR, reflecting the intentions of the algorithm, as shown in the results of Fig. 7b.

Second, Fig. 8 demonstrates the significant advantage of the TGRF in terms of learning speed compared to the reward function without the TGRF. As shown in Fig. 8a, the three models reached saturation after 6000 episodes. At this point, the model applying TGRF achieved the highest SR. This indicates that the TGRF contributes to faster learning speeds.

However, the TGRF has limitations in crowded environments. It does not inherently enhance the performance of the core algorithms. Comparable performance was achieved for certain policies, as shown in Tables I and II. This suggests that the TGRF expedites the algorithm to achieve optimal performance, rather than enhancing the algorithm itself.

\section{Real-World Experiments}\label{sec5}

\begin{figure}[h]%
\centering
\includegraphics[width=1.0\textwidth]{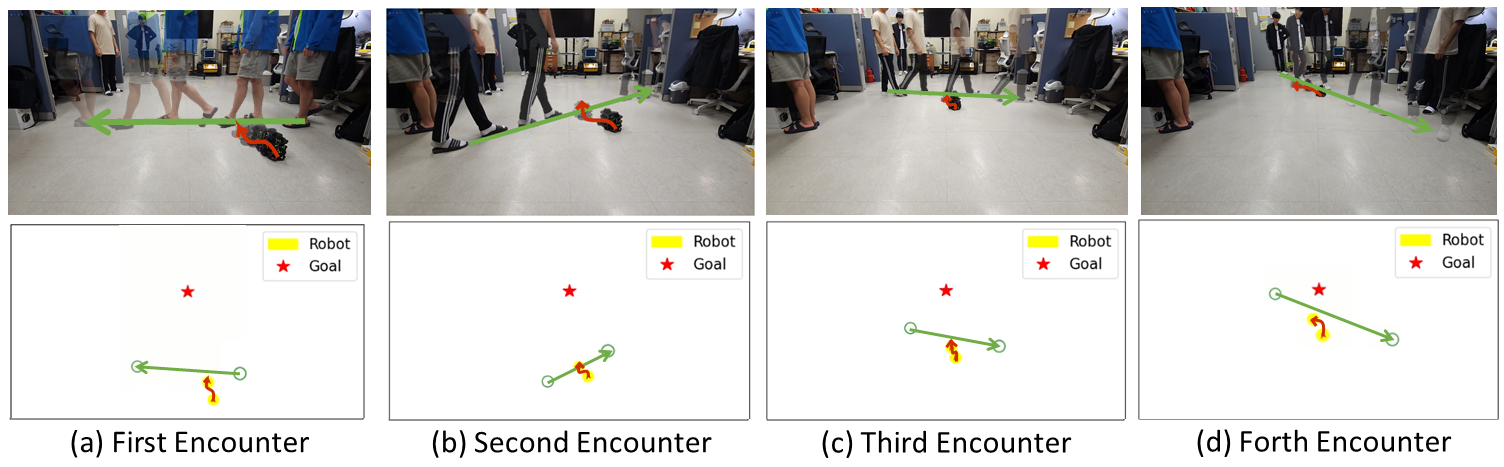}
\caption{Evasive actions performed by robot in real-world scenarios and corresponding renderings with four humans (from left to right). Green arrows represent the movement path of human, while red arrows indicate the movement path of robot. In addition, yellow circles indicate robots, green circles represent humans, and red stars indicate destinations. These illustrations showcase avoidance strategies employed by the robot as it encounters successive individuals: (a) first, (b) second, (c) third, and (d) fourth human.}\label{fig9.png}
\end{figure}

This study extended beyond simulations to real-world experiments. The model trained using the unicycle approach was applied to a physical robot in a real environment. Unicycle kinematics presents limitations in terms of direction and position control when compared to holonomic kinematics. Our experimental setup consisted of a host computer equipped with an Intel i5-10600 processor at 3.30 GHz and an NVIDIA RTX 3070 GPU, which was integrated with a Turtle-bot3. A LiDAR sensor, LDS-01, played a pivotal role in human detection and robot position estimation. Robot positioning relies on SLAM localization, and human detection is accomplished using a 2D people detection algorithm based on 2D LiDAR data [35].

Although our study assumed the absence of static obstacles other than humans, our real-world experiments were conducted in a confined space measuring approximately 3 × 5m with static obstacles. These experiments involved scenarios in which the robot navigated between predefined start and destination points and encountered one to four pedestrians along its path. The maximum speed was approximately 0.6 m/s, and the investigation covered scenarios involving four moving individuals.

As shown in Fig. 9a, the robot faces diagonally upward as the pedestrian moves from right to left. In this scenario, the robot rotated to the left, aligning with the pedestrian’s direction of movement, instead of moving behind (to the right) the human. This decision appears rational because both the destination and the robot’s current orientation are oriented diagonally upward, making a leftward maneuver the most efficient choice when considering the human direction, speed, and destination. Notably, in another experiment involving three individuals, the robot was observed halting temporarily instead of moving to the left.

In Fig. 9b, the robot encounters a pedestrian walking diagonally from left to right. In response, the robot navigated to the left to avoid obstructing the path of the pedestrian.

In Fig. 9c, the robot faces a human walking from left to right. Similarly, it predicts the human's trajectory and executes a leftward turn to avoid collisions while approaching the destination.

In Fig. 9d, the robot encounters a human crossing diagonally from left to right near the destination. The robot smartly avoided humans by initially turning left, avoiding the pedestrian, and then turning right to reach its destination.

These actions underscore the robot’s ability to make real-time decisions based on a dynamic environment, considering factors such as human path, velocity, and proximity to the destination. The robot’s avoidance strategies prioritize efficiency while maintaining safety and are influenced by various factors, including its current orientation and the overall context of the situation. These real-world experiments verified the adaptability of the model in complex and dynamic environments, where human–robot interactions demand responsive and context-aware behavior. Comprehensive renderings and additional experimental videos are available on https://youtu.be/9x24k75Zj5k?si=OtczdVXPUnbGwpv-.

Two primary limitations were encountered during this experiment. First, the computation load: The use of DNNs for action and trajectory predictions significantly increases the computational demands. Considering the number of pedestrians, particularly for trajectory prediction, the time required for the next action was approximately 0.22 s. This resulted in irregular robot movements and delayed pedestrian responses. Second, physical constraints: The accuracy of human detection and prediction is affected by sensor noise, limitations in detection performance, and challenges in determining human angles. These factors lead to occasional misidentification of obstacles as humans or limitations in the precision of human location information, thereby reducing the accuracy of trajectory prediction. In addition, noise from the LiDAR sensor and location information errors caused by the movement of the robot accumulated over time, resulting in inaccuracies in the location values as the experiment progressed.

\section{Results and Future Research}\label{sec6}

This paper introduces a TGRF specifically designed for robots navigating crowded environments. The TGRF offers several advantages, including high performance with minimal hyperparameters, adaptability to diverse objectives, and expedited learning and stabilization processes. These claims are supported by the success rates achieved and the algorithm’s enhanced ability to discern human intentions when compared to previous reward functions.

However, challenges have emerged in both the simulations and real-world experiments. In the simulations, these challenges involved sensitivity to hyperparameters, algorithmic limitations, a trade-off correlation between SR and NT, and the absence of static obstacles. In real-world tests, the challenges include sensor noise and physical constraints.

Hence, in future research, we propose two key strategies, (1) we will apply the TGRF to various environments and with different objects. While our study demonstrates its effectiveness primarily with human rewards, we plan to expand our experiments by applying the TGRF to diverse objects, such as walls, obstacles, and drones. (2) we will devise a TGRF that considers physical limitations. Although the TGRF performs exceptionally well under ideal conditions, its perfor-mance decreases in reality due to computational load and physical issues. Therefore, we aim to implement a dynamically adaptive TGRF that adjusts according to the situation by incorporating knowledge regarding these physical limitations.

\section*{Funding}
This research was partly supported by the MSIT(Ministry of Science and ICT), Korea, under the Convergence security core talent training business support program(IITP-2023-RS-2023-00266615) supervised by the IITP(Institute for Information \& Communications Technology Planning \& Evaluation), and the BK21 plus program "AgeTech-Service Convergence Major" through the National Research Foundation (NRF) funded by the Ministry of Education of Korea[5120200313836], and Institute of Information \& communications Technology Planning \& Evaluation (IITP) grant funded by the Korea government(MSIT) (No.RS-2022-00155911, Artificial Intelligence Convergence Innovation Human Resources Development (Kyung Hee University)), and the Ministry of Trade, Industry and Energy (MOTIE), South Korea, under Industrial Technology Innovation Program under Grant 20015440, 20025094

\section*{References}

\author{}
\date{}
\maketitle

\begin{enumerate}
    \item Nourbakhsh, I.R., et al., Mobile robot obstacle avoidance via depth from focus. Robotics and Autonomous Systems, 1997. 22(2): p. 151-158.
    
    \item Ulrich, I. and J. Borenstein. VFH+: Reliable obstacle avoidance for fast mobile robots. in Proceedings. 1998 IEEE international conference on robotics and automation (Cat. No. 98CH36146). 1998. IEEE.
    
    \item Nalpantidis, L. and A. Gasteratos, Stereovision-based fuzzy obstacle avoidance method. International Journal of Humanoid Robotics, 2011. 8(01): p. 169-183.
    
    \item Nalpantidis, L., G.C. Sirakoulis, and A. Gasteratos, Non-probabilistic cellular automata-enhanced stereo vision simultaneous localization and mapping. Measurement Science and Technology, 2011. 22(11): p. 114027.
    
    \item Pritsker, A.A.B., Introduction to Simulation and SLAM II. 1984: Halsted Press.
    
    \item Grisetti, G., et al., A tutorial on graph-based SLAM. IEEE Intelligent Transportation Systems Magazine, 2010. 2(4): p. 31-43.
    
    \item Ai, Y., et al., DDL-SLAM: A robust RGB-D SLAM in dynamic environments combined with deep learning. IEEE Access, 2020. 8: p. 162335-162342.
    
    \item Cui, L. and C. Ma, SDF-SLAM: Semantic depth filter SLAM for dynamic environments. IEEE Access, 2020. 8: p. 95301-95311.
    
    \item Borenstein, J. and Y. Koren, Real-time obstacle avoidance for fast mobile robots. IEEE Transactions on systems, Man, and Cybernetics, 1989. 19(5): p. 1179-1187.
    
    \item Van Den Berg, J., et al. Reciprocal n-body collision avoidance. in Robotics Research: The 14th International Symposium ISRR. 2011. Springer.
    
    \item Helbing, D. and P. Molnar, Social force model for pedestrian dynamics. Physical review E, 1995. 51(5): p. 4282.
    
    \item Patel, U., et al., DWA-RL: Dynamically Feasible Deep Reinforcement Learning Policy for Robot Navigation among Mobile Obstacles, in 2021 IEEE International Conference on Robotics and Automation (ICRA). 2021. p. 6057-6063.
    
    \item Shuijing Liu, P.C., Zhe Huang, Neeloy Chakraborty, Kaiwen Hong, and D.L.M. Weihang Liang, Junyi Geng, and Kathe-rine Driggs-Campbell. Intention Aware Robot Crowd Navigation with Attention-Based Interaction Graph. 2023.
    
    \item Chen, C., et al. Crowd-robot interaction: Crowd-aware robot navigation with attention-based deep reinforcement learning. in 2019 international conference on robotics and automation (ICRA). 2019. IEEE.
    
    \item Van den Berg, J., M. Lin, and D. Manocha. Reciprocal velocity obstacles for real-time multi-agent navigation. in 2008 IEEE international conference on robotics and automation. 2008. Ieee.
    
    \item Oh, J., et al., SCAN: Socially-Aware Navigation Using Monte Carlo Tree Search. 2022.
    
    \item Liu, S., et al. Decentralized structural-rnn for robot crowd navigation with deep reinforcement learning. in 2021 IEEE International Conference on Robotics and Automation (ICRA). 2021. IEEE.
    
    \item Kretzschmar, H., G. Grisetti, and C. Stachniss, Lifelong map learning for graph-based slam in static environments. KI-Künstliche Intelligenz, 2010. 24: p. 199-206.
    
    \item Brown, N., Edward T. Hall: Proxemic Theory, 1966. Center for Spatially Integrated Social Science. University of Cali-fornia, Santa Barbara. http://www.csiss.org/classics/content/13 Read, 2001. 18: p. 2007.
    
    \item Rios-Martinez, J., A. Spalanzani, and C. Laugier, From proxemics theory to socially-aware navigation: A survey. Inter-national Journal of Social Robotics, 2015. 7: p. 137-153.
    
    \item Bellman, R., A Markovian decision process. Journal of mathematics and mechanics, 1957: p. 679-684.
    
    \item Hastings, W.K., Monte Carlo sampling methods using Markov chains and their applications. 1970.
    
    \item Mnih, V., et al., Playing atari with deep reinforcement learning. arXiv preprint arXiv:1312.5602, 2013.
    
    \item Sutton, R.S., Learning to predict by the methods of temporal differences. Machine learning, 1988. 3: p. 9-44.
    
    \item Jeong, H., et al., Deep Reinforcement Learning for Active Target Tracking, in 2021 IEEE International Conference on Robotics and Automation (ICRA). 2021. p. 1825-1831.
    
    \item Gleave, A., et al., Quantifying differences in reward functions. arXiv preprint arXiv:2006.13900, 2020.
    
    \item Mataric, M.J., Reward functions for accelerated learning, in Machine learning proceedings 1994. 1994, Elsevier. p. 181-189.
    
    \item Laud, A.D., Theory and application of reward shaping in reinforcement learning. 2004: University of Illinois at Urba-na-Champaign.
    
    \item Montero, E.E., et al., Dynamic warning zone and a short-distance goal for autonomous robot navigation using deep reinforcement learning. Complex \& Intelligent Systems, 2023: p. 1-18.
    
    \item Samsani, S.S. and M.S. Muhammad, Socially compliant robot navigation in crowded environment by human behavior resemblance using deep reinforcement learning. IEEE Robotics and Automation Letters, 2021. 6(3): p. 5223-5230.
    
    \item Samsani, S.S., H. Mutahira, and M.S. Muhammad, Memory-based crowd-aware robot navigation using deep rein-forcement learning. Complex \& Intelligent Systems, 2023. 9(2): p. 2147-2158.
    
    \item Choi, J., G. Lee, and C. Lee, Reinforcement learning-based dynamic obstacle avoidance and integration of path plan-ning. Intelligent Service Robotics, 2021. 14: p. 663-677.
    
    \item Liu, S., et al., Socially aware robot crowd navigation with interaction graphs and human trajectory prediction. arXiv preprint arXiv:2203.01821, 2022.
    
    \item Pérez-D’Arpino, C., et al. Robot navigation in constrained pedestrian environments using reinforcement learning. in 2021 IEEE International Conference on Robotics and Automation (ICRA). 2021. IEEE.
    
    \item Scholz, J., et al. Navigation Among Movable Obstacles with learned dynamic constraints. in 2016 IEEE/RSJ Interna-tional Conference on Intelligent Robots and Systems (IROS). 2016. IEEE.
    
    \item Cassandra, A.R. A survey of POMDP applications. in Working notes of AAAI 1998 fall symposium on planning with partially observable Markov decision processes. 1998.
    
    \item Hu, Y., et al., Learning to utilize shaping rewards: A new approach of reward shaping. Advances in Neural Infor-mation Processing Systems, 2020. 33: p. 15931-15941.
    
    \item Icarte, R.T., et al., Reward machines: Exploiting reward function structure in reinforcement learning. Journal of Arti-ficial Intelligence Research, 2022. 73: p. 173-208.
    
    \item Yuan, M., et al., Automatic Intrinsic Reward Shaping for Exploration in Deep Reinforcement Learning. arXiv preprint arXiv:2301.10886, 2023.
    
    \item Zhang, S., et al. Average-reward off-policy policy evaluation with function approximation. in international confer-ence on machine learning. 2021. PMLR.
    
    \item Rucker, M.A., et al., Reward Shaping for Human Learning via Inverse Reinforcement Learning. arXiv preprint arXiv:2002.10904, 2020.
    
    \item Goyal, P., S. Niekum, and R.J. Mooney, Using natural language for reward shaping in reinforcement learning. arXiv preprint arXiv:1903.02020, 2019.
    
    \item Trautman, P. and A. Krause. Unfreezing the robot: Navigation in dense, interacting crowds. in 2010 IEEE/RSJ Interna-tional Conference on Intelligent Robots and Systems. 2010. IEEE.
    
    \item Huang, Z., et al., Learning sparse interaction graphs of partially detected pedestrians for trajectory prediction. IEEE Robotics and Automation Letters, 2021. 7(2): p. 1198-1205.
    
    \item Niu, Z., G. Zhong, and H. Yu, A review on the attention mechanism of deep learning. Neurocomputing, 2021. 452: p. 48-62.
    
    \item Fu, R., Z. Zhang, and L. Li. Using LSTM and GRU neural network methods for traffic flow prediction. in 2016 31st Youth academic annual conference of Chinese association of automation (YAC). 2016. IEEE.
    
    \item Goodman, N.R., Statistical analysis based on a certain multivariate complex Gaussian distribution (an introduction). The Annals of mathematical statistics, 1963. 34(1): p. 152-177.
    
\end{enumerate}

\end{document}